\documentclass[10pt,twocolumn,letterpaper]{article}

\usepackage{cvpr}       % To produce the CAMERA-READY version
\usepackage[dvipsnames]{xcolor}

\usepackage{url}
\usepackage{multirow}
\usepackage{multicol}
\definecolor{cvprblue}{rgb}{0.21,0.49,0.74}
\usepackage[pagebackref,breaklinks,colorlinks,citecolor=cvprblue]{hyperref}

\def\paperID{11135} % *** Enter the Paper ID here
\def\confName{CVPR}
\def\confYear{2024}

\title{A Speed Odyssey for Deployable Quantization of LLMs}

\author{Qingyuan Li, Ran Meng, Yiduo Li, Bo Zhang, Liang Li, Yifan Lu, Xiangxiang Chu, Yerui Sun, Yuchen Xie\\
Meituan Inc. \\
\small \texttt{\{liqingyuan02,mengran03,liyiduo,zhangbo97,liliang58\}@meituan.com} \\
}

\begin{document}
\maketitle

\begin{abstract}
The large language model era urges faster and less costly inference. Prior model compression works on LLMs tend to undertake a software-centric approach primarily focused on the simulated quantization performance. By neglecting the feasibility of deployment, these approaches are typically disabled in real practice. They used to drastically push down the quantization bit range for a reduced computation which might not be supported by the mainstream hardware, or involve sophisticated algorithms that introduce extra computation or memory access overhead. We argue that pursuing a hardware-centric approach in the construction of quantization algorithms is crucial. In this regard, we are driven to build our compression method on top of hardware awareness, eliminating impractical algorithm choices while maximizing the benefit of hardware acceleration. Our method, OdysseyLLM, comes with a novel W4A8 kernel implementation called FastGEMM and a combined recipe of quantization strategies. Extensive experiments manifest the superiority of our W4A8 method which brings the actual speed boosting up to \textbf{4$\times$} compared to Hugging Face~\footnote{\url{https://github.com/huggingface/transformers}} FP16 inference and \textbf{2.23$\times$} vs. the state-of-the-art inference engine TensorRT-LLM~\footnote{\url{https://github.com/NVIDIA/TensorRT-LLM}} in FP16, and \textbf{1.45$\times$} vs. TensorRT-LLM in INT8, yet without substantially harming the performance.
\end{abstract}

\section{Introduction}\label{sec:intro}

\begin{figure}[h]
\centering
\includegraphics[width=0.4\textwidth]{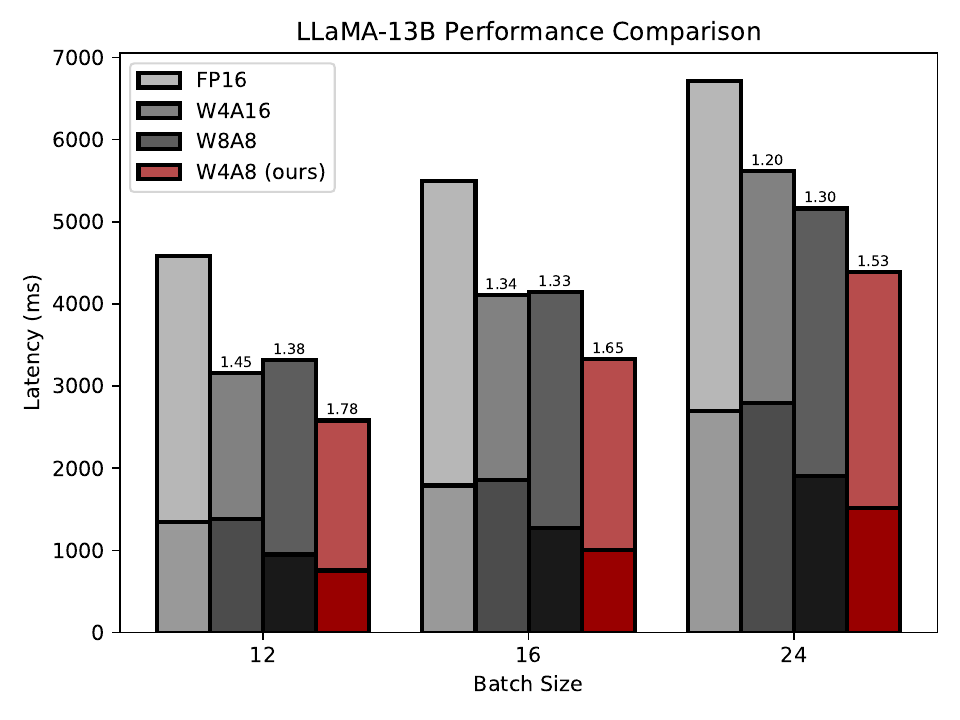}
\caption{Inference latency on LLaMA-13B quantized in various bit widths. Tested under an input of 1024 tokens and an output of 128 tokens with tensor parallelism on a single A100-80G GPU. All implementations share the same techniques to have a fair comparison. The lower half of a bar exhibits the context decoding stage and the higher half shows the self-decoding stage.}
\vskip -0.2in
\label{fig:perf}
\end{figure}

Large language models (LLMs) such as GLM \cite{du2021glm}, BLOOM \cite{laurencconbigscience}, OPT \cite{zhang2022opt} and LLaMA series \cite{touvron2023llama,touvron2023llama2} possess the powerful ability of ``emergent knowledge" and have revolutionized the field of natural language processing, which opens up a new era for artificial intelligence. However, the massive scale of these models requires enormous storage and computational resources, posing a series of challenges for deployment even for industrial high-end server GPUs, let alone mobile or edge computing devices where computational resource limitations tremendously hinder the widespread application of these models.

%%%%%%%%%%%%% MOVED FOR LAYOUT %%%%%%%%%%%
\begin{figure*}[ht]
\centering
\includegraphics[width=0.9\textwidth]{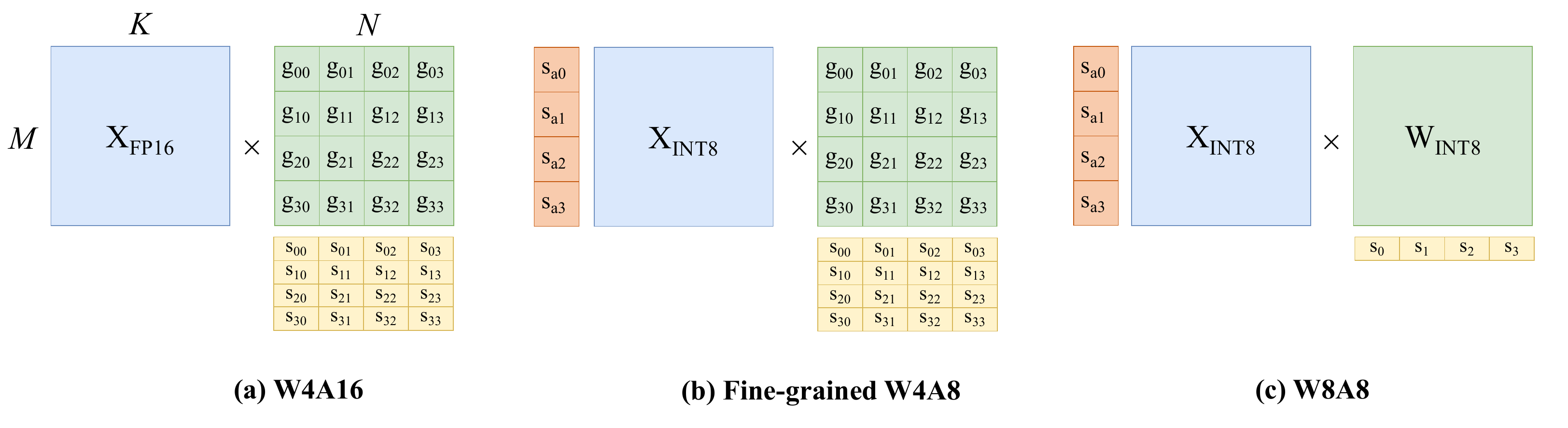}
\caption{Comparison of current MatMul paradigm in practical bit widths.  $X$ is the input (activation) of shape $M \times K$, and $g_{00}$ denotes a group of weights, $s_{a}$ is the scale factor for activation, and $s$ is the scale for weights.} %We choose W4A8 and remove the groupwise scaling of weights.
\label{fig:bitwidth-comp}
\end{figure*}
%%%%%%%%%%%%%%%%%%%%%%%%%%%%%%%%%%%

Compressing large models is a highly challenging task. Current methods generally fall into the categories of \emph{pruning}, \emph{distillation}, \emph{quantization}, and \emph{low-rank decomposition}. To name a few, LLM-Pruner~\cite{ma2023llmpruner} is a structured pruning method, where the pruning rate is limited to ~20\% to retain the model capability, and the pruned model requires further fine-tuning to restore accuracy. SparseGPT~\cite{frantar2023sparsegpt} and Wanda~\cite{sun2023simple} compress large models using unstructured sparsity. Although re-training is not required, they are heavily hardware-dependent and have very limited use. On the contrary, quantization is a more universal compression method, and most hardware provides dedicated computation acceleration units for integers. However, as the scale of large models increases, the outlier phenomenon in activation values becomes more severe. Traditional quantization methods for activation values can lead to significant quantization errors. SmoothQuant~\cite{xiao2023smoothquant} achieves almost lossless W8A8 quantization (Fig.~\ref{fig:bitwidth-comp} (c)) by migrating the quantization difficulty of activation values to weights. To further reduce the cost, GPTQ~\cite{frantar2022gptq} quantizes model weights to INT4, avoiding the quantization problem of activation values and achieving significant performance gains. Unfortunately, the fine-granularity (Fig.~\ref{fig:bitwidth-comp} (a)) in its solution inevitably creates overhead that cancels out the lower-bit benefits. A vanilla adaption of GPTQ to have a fine-grained W4A8 strategy as in Fig.~\ref{fig:bitwidth-comp} (b) naturally inherits the disadvantages, let alone the performance degradation when we chase lower-bit quantization. Therefore, designing a practical W4A8 quantization scheme is urgent but it requires a deeper and thorough rethinking.

%In this paper, we propose a W4A8 quantization method that combines the advantages of previous methods. We quantize activations using 8-bit quantization, weights using 4-bit quantization, and matrix calculations using 8-bit quantization. This approach can accelerate actual computation processes and improve I/O utilization. Additionally, our method can stack KV Cache INT8 quantization, further saving memory resources. Most importantly, we introduce a PTQ scheme that does not require fine-tuning for training, enabling rapid production of quantized models. We experiment with multiple large models and find that our method is nearly lossless.

In this paper, we are driven to invert the common practice by taking in a hardware-centric approach. We seek real deployment that requires \emph{reduced memory footprint}, \emph{boosted inference speed}, and \emph{non-degraded quantized performance}. We argue that this shift is crucial in developing new algorithms so that the outcome is readily applicable. Hardware constraints help us eliminate the impractical choices to have a reduced space of trial-and-error, shedding light on a viable solution in the meantime.  
In a nutshell, our contributions can be summarized as follows,

\begin{enumerate}
\item We advocate a hardware-centric approach that ultimately leads to a deployable solution which is a crucial paradigm change in algorithm construction.
\item We provide the first deployable W4A8 solution, codenamed as OdysseyLLM, that comprises a tailored quantization configuration and a novel FastGEMM kernel for 4-bit integer matrix multiplication that dramatically reduces the cost, and it achieves \textbf{2.23$\times$} and \textbf{1.45$\times$} speed boosting over the TensorRT-LLM FP16 and INT8 implementation respectively.
\item Our W4A8 recipe is proven mostly on par with the state-of-the-art W8A8 quantization method SmoothQuant on a variety of common language benchmarks for the state-of-the-art LLMs.
\end{enumerate}

\section{Related Work}\label{sec:rw}
\textbf{Taxonomy of LLM Compression.} A recent thorough survey \cite{zhu2023survey} categorizes LLM compression methods into pruning~\cite{frantar2023sparsegpt,sun2023simple,zhang2023pruning}, knowledge distillation~\cite{gu2023knowledge}, quantization~\cite{xiao2023smoothquant,frantar2022gptq}, low-rank factorization~\cite{zhang2023pruning,wu2023zeroquantfp}. While each category shows promising gains, in this paper we primarily focus on low-bit quantization for extreme speed boosting. The orthogonal composition of these methods is also tempting, \eg, LoRAPrune~\cite{zhang2023pruning} obtains a 50\% compression ratio with structured pruning combined by LoRA~\cite{hu2021lora}. QLoRA~\cite{dettmers2023qlora} efficiently finetunes the 4-bit quantized LLMs with low-rank adaptation ~\cite{hu2021lora} as well. Nevertheless, there is still room for further investigation. 

\noindent \textbf{Variants of Bit Widths in Quantization.}
Common choices of bit widths for LLM quantization are W8A8~\cite{xiao2023smoothquant,yao2022zeroquant}, W4A16~\cite{frantar2022gptq,lin2023awq,shao2023omniquant}, W4A8~\cite{yao2023zeroquantv2,li2023fptq}, and W4A4~\cite{yuan2023rptq}. W8A8 suffers from limited acceleration for token generation while W4A16 is relatively slow during the pre-filling stage. Current W4A8 recipes usually adopt a fine-grained strategy that hampers the inference. Going further with W4A4 harms the performance and also induces complexity for implementation. There are a few mixed-precision quantization methods like LLM.int8()~\cite{dettmers2022llm} and QUIK~\cite{ashkboos2023towards}, where the calculation of some outlier layers fallback in FP16 as a trade-off between accuracy and latency. Besides, quantization that exploits the low-bit floating point representation (FP4, FP8) is analogous to non-uniform integer quantization, which exhibits improved performance as well~\cite{wu2023zeroquantfp,liu2023llmfp4}. However, they are either restricted to certain GPUs only or no such hardware is available yet.

\section{Preliminary Knowledge on Quantization}~\label{sec:prelim}

We show a glossary of common LLM quantization terms and techniques that build up the recipes of current state-of-the-art compression methods.

\paragraph{Weight-only vs. Weight and Activation} Quantization methods vary on whether only the weights $\mathbf{W}$ are quantized (weight-only) \cite{frantar2022gptq} or both the weights $\mathbf{W}$ and the activations $\mathbf{X}$ are quantized \cite{xiao2023smoothquant}. The latter is more complicated and possibly induces more quantization loss but it will be worth the pain for a decreased model size and an increased inference speed.

\paragraph{Layerwise Quantization} This is the most common scheme that iteratively quantizes each layer to obtain $\mathbf{W}_{\mathbf{q}}$ by minimizing the mean square error before and after quantization. For instance, when combined with the weight-only strategy, each layer has to solve Eq.~\ref{eq:layerwise}. 

\begin{align}\label{eq:layerwise}
\operatorname{argmin}_{\mathbf{w}} \vert \mathbf{W}\mathbf{X} - \mathbf{W}_{\mathbf{q}} \mathbf{X} \vert_2^2
\end{align}

\paragraph{Symmetric vs. asymmetric} Eq.~\ref{eq:asym} shows the quantization $\mathbf{Q}$ and dequantization process $\mathbf{D}$ respectively. Symmetric quantization sets the zero point $z$ as 0. While asymmetric quantization needs to find the optimal position for the zero point to minimize the quantization error. This however incurs additional subtraction and possible overflow. 
\begin{align}\label{eq:asym}
\mathbf{Q}(x) &= round((x - z) / scale) \\
\mathbf{D}(\mathbf{Q}(x)) &= \mathbf{Q}(x) * scale + z
\end{align}

\paragraph{Per channel vs. fine-grained} Per channel quantization keeps a quantization scale for each channel, while in fine-grained quantization (also known as group-wise or per group), weight channels are further assembled into groups that have a more complex representation. Fig.~\ref{fig:bitwidth-comp} well illustrates their difference. Due to the intricate computing pipeline (Eq.~\ref{eq:gemm2}), the fine-grained method inevitably prolongs the inference time.

\paragraph{Per tensor vs. Per token} For activation quantization, it is advisable to adopt a per-token strategy to improve the performance over per tensor strategy (Fig.~\ref{fig:bitwidth-comp}), where each activation token corresponds to a quantization scale, also shown in Fig.~\ref{fig:bitwidth-comp} (b) and (c). However, per token will increase a moderate amount of overhead.

%%%%% MOVED FOR LAYOUT %%%%%%%%%%%%

\begin{table*}[ht]
\centering
\setlength{\tabcolsep}{1pt} 
\begin{tabular}{l*{8}{c}}
\toprule
Method	&	Bits	&	Granularity	&	LLaMA-1-7B	&	LLaMA-1-13B	&	LLaMA-1-65B	&	LLaMA-2-7B	&	LLaMA-2-13B	&	LLaMA-2-70B \\
\midrule
FP16	&	W16A16	&	None	&	73.74\%	&	76.19\%	&	79.20\%	&	73.70\%	&	76.64\%	&	79.57\% \\
RTN\textsubscript{pt}	&	W16A8	&	None	&	73.26\% \textsubscript{-0.5\%}	&	76.13\% \textsubscript{(-0.1\%)}	&	78.67\% \textsubscript{(-0.5\%)}	&	73.70\%	&	76.83\% \textsubscript{(+0.2\%)}	& 79.12\% \textsubscript{(-0.5\%)}	\\
RTN\textsubscript{g128}	&	W4A16	&	g128	&	72.87\% \textsubscript{(-0.9\%)}	&	75.08\% \textsubscript{(-1.1\%)}	&	78.87\% \textsubscript{(-0.3\%)}	&	71.24\% \textsubscript{(-2.5\%)}	&	76.29\% \textsubscript{(-0.4\%)}	&	78.69\% \textsubscript{(-0.9\%)} \\
GPTQ\textsubscript{g128}	&	W4A16	&	g128	&	70.21\% \textsubscript{(-3.5\%)}	&	75.68\% \textsubscript{(-0.5\%)}	&	78.77\% \textsubscript{(-0.5\%)}	&	72.31\% \textsubscript{(-1.4\%)}	&	75.99\% \textsubscript{(-0.7\%)}	&	79.86\% \textsubscript{(-0.3\%)} \\
RTN	&	W4A16	&	pc	&	65.34\% \textsubscript{(-8.4\%)}	&	69.42\% \textsubscript{(-6.8\%)}	&	75.63\% \textsubscript{(-3.6\%)}	&	64.25\% \textsubscript{(-9.5\%)}	&	73.08\% \textsubscript{(-3.6\%)}	&	77.02\% \textsubscript{(-2.6\%)} \\
GPTQ\textsubscript{ro}	&	W4A16	&	pc	&	67.92\% \textsubscript{(-5.8\%)}	& 71.05\% \textsubscript{(-6.0\%)}		&	77.12\% \textsubscript{(-2.1\%)}	&	68.95\% \textsubscript{(-4.8\%)}	&	74.35\% \textsubscript{(-2.3\%)}	&	78.83\% \textsubscript{(-0.7\%)} \\
 \bottomrule
 \end{tabular}
 \caption{Accuracy comparison of different quantization methods on the LAMBADA dataset for the LLaMA series models. pt: per-token, pc: per-channel, g128: 128 groups, ro: with activation reordering}
 \label{tab:trade-off-lambada}
\end{table*}

%%%%%%%%%%%%%%%%%%%

\section{Motivation}\label{sec:mot}

Albeit the advances in LLM compression, the inference still incurs high latencies due to its intrinsically low parallelizability and huge memory footprints. Facing two main challenges below, we are driven to scheme a specific method to continue exploring the possible upper limits.
\subsection{Architecture Limitations}\label{subsec:arch}
 Large language models adopt the Transformer structure~\cite{vaswani2017attention} and generate tokens in a ``self-regressive" manner. The entire inference process can be divided into the \emph{context decoding} stage (pre-filling) and the \emph{self-decoding} stage (token generation). Context decoding stage under long input conditions is a typical computationally intensive task. The self-decoding stage instead, due to the non-parallel limitations of the ``self-regressive" generation, cannot effectively utilize hardware resources, rendering it a typical memory-intensive task. This is also depicted in the roofline analysis in ~\cite{ashkboos2023towards}. Among the up-to-date quantization methods, the W8A8 recipe has a significant acceleration effect in the context decoding stage, while W4A16, due to its ability to further reduce memory bandwidth, has more advantages in the self-decoding stage. Theoretically, we can have W4A8 that combines these two quantization methods to enjoy the optimization benefits of both two stages. However, this direction is largely stalled by degraded performance and hardware constraints to be discussed below.

\subsection{Hardware Constraints}\label{subsec:hard-cons}

Current mainstream INT4 schemes~\cite{frantar2022gptq,li2023fptq} have widely adopted fine-grained quantization to counter the degradation effect. In Fig.~\ref{fig:bitwidth-comp}, we investigate the implementation of three popular quantization recipes of different bit widths to exhibit the drawbacks of such choices. W4A16 (Fig.~\ref{fig:bitwidth-comp}a) performs 4-bit groupwise quantization on weights. During matrix multiplication, it is costly to dequantize ($\mathbf{Dq}$) the INT4 weights into FP16 in real-time before the actual calculation, as described in Eq.~\ref{eq:gemm1}. W4A8 (Fig.~\ref{fig:bitwidth-comp}b) performs 4-bit groupwise quantization on weights and uses 8-bit per-token quantization for activation. Similarly, it needs to first convert the INT4 weights into INT8 before GEMM operations, formulated in Eq.~\ref{eq:gemm2}. Due to the use of group-wise quantization, each group has to be dequantized back to FP32 when accumulating, which brings considerable overhead. For W8A8 (Fig.~\ref{fig:bitwidth-comp}c) specified in Eq.~\ref{eq:gemm3} and Eq.~\ref{eq:gemm4}, it uses 8-bit per-channel quantization for weights and 8-bit per-token quantization for activation. The dequantization is performed after GEMM, which is so far the most hardware-friendly process. 
%This observation pushes us to design a non-fine-grained W4A8 solution, but with certain tactics to resolve degradation.

%Let $W_{g,j}$ be the weights of group $g$ at column $j$, $S_{g,j}$ be the quantization scale for $W_{g,j}$, $A_{i,g}$ be the input for this group. To perform MatMul in the W4A16 paradigm, we have to first dequantize weights by group into FP16 and have them multiplied by $A_{i,g}$. The accumulator is typically in a bit width of 32 and thus we have an FP32 output.

\begin{align}\label{eq:gemm1}
O_{i,j}^{FP32} &= \sum_g (A_{i,g} \times \mathbf{Dq}(\mathbf{W}_{g,j}^{\top}, S_{g,j}) ) \\
\label{eq:gemm2}
O_{i,j}^{FP32} &= \sum_g ( \mathbf{Dq}(A_{i,g} \times \texttt{UINT4toS8}(\mathbf{W}_{g,j}^{\top}), S_{a_i}, S_{g,j}) \\
\label{eq:gemm3}
O_{i,j}^{FP32} &= \mathbf{Dq}( \sum_k (A_{i,k} \times \mathbf{W}_{k,j}^{\top}), S_{i,j}) \\
\label{eq:gemm4}
S_{i,j} &= S_{a_i} \cdot S_{w_j}
\end{align}

\textbf{Trade-off between precision and speed?} In Table~\ref{tab:trade-off-lambada}, we first test the performance of the LLaMA models~\cite{touvron2023llama} with various quantization methods on the LAMBADA~\cite{paperno2016lambada} dataset. If we only quantize activations, RTN-pt readily delivers results close to FP16, which prevents us from resorting to activation smoothing methods~\cite{xiao2023smoothquant,li2023fptq}. For fine-grained quantization methods for weights, both vanilla Round-To-Nearest (RTN-g128) and GPTQ-g128~\cite{frantar2022gptq} can retain model accuracy. However, after adopting per-channel quantization, RTN has an accuracy drop in the range of 3\% to 10\%, while GPTQ with a reordering trick (higher error-prone channels are quantized first) has 1\% to 6\%.

As mentioned in Sec~\ref{subsec:hard-cons}, fine-grained W4A8 requires a large number of \texttt{Dequantize} operations to be inserted in the GEMM calculation process, bringing non-negligible additional overhead, thereby offsetting the speed advantage brought by INT8 GEMM. It seems impossible to achieve both accuracy and speed. 

To fulfill our goal, we have no other choice but to abandon the fine-grained weight quantization strategy and pursue the per-channel weight quantization instead. To compensate for the caused accuracy loss therein, we are forced to involve particular quantization schemes to make the performance comparable to that of fine-grained ones. Additionally, to utilize hardware resources in full, we are compelled to rewrite a specific GEMM operation for W4A8, which we later call FastGEMM. 
\section{OdysseyLLM}\label{sec:method}

Our proposed method, codenamed OdysseyLLM, records the way to a viable W4A8 solution.

\subsection{Adaptive Weight Clipping}\label{subsec:adapt}
The weights of neural networks generally exhibit a Gaussian distribution. As INT4 has only half the bit widths of INT8, a typical min-max uniform 4-bit quantization method~\cite{wu2020integer,liu2023llm} will cause a large number of rounding errors, especially near the weights close to 0, leading to a detrimental quantization degradation. To alleviate this phenomenon, one can resort to clipping the range of weights. For example, LSQ~\cite{esser2019learned} and PACT~\cite{choi2018pact} adaptively learn a truncation value of the weight. However, the direct learning truncation value method does not have obvious benefits in low-bit quantization on LLMs.

In this regard, Ominiquant~\cite{shao2023omniquant} proposed \emph{Learnable Weight Clipping}, which learns the truncation intensity (denoted by $\gamma$ and $\beta$) of each channel. The optimal truncation value is obtained by optimizing it through the gradient descent method. Motivated by the hardware-centric principle, we revise their approach into a symmetric version (Eq.~\ref{eq:clamp}) as it is more hardware-efficient.  

\begin{align}\label{eq:clamp}
\mathbf{W}_{\mathbf{q}}&=\operatorname{clamp}\left(\left\lfloor\frac{\mathbf{W}}{S}\right\rceil, -2^{N-1},2^{N-1}-1\right) \\ 
S&=\frac{\max(\vert \gamma \max (\mathbf{W})\vert, \vert \beta \min (\mathbf{W})\vert)}{2^{N-1}-1} %\\
%z&=-\left\lfloor\frac{\beta \min (\mathbf{W})}{s}\right\rceil
\end{align}

\begin{figure}[ht]
\centering
\includegraphics[width=0.45\textwidth]{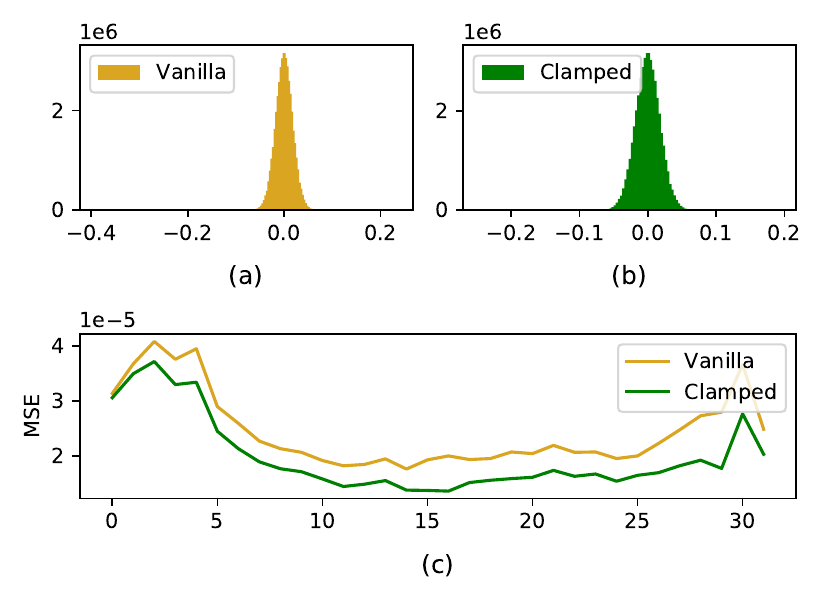}
\caption{\textbf{Top:} Weight distribution of vanilla weights (a) and clamped weights (b). \textbf{Bottom:} Comparison of layerwise MSE of per-channel fake quantization with clamped and vanilla weights.}
\label{fig:layerwise-mse}
\end{figure}

We finally obtain a more compact weight distribution, \eg, $(-0.4, 0.2)$ narrowed to $(-0.2, 0.2)$, as shown in the Fig.~\ref{fig:layerwise-mse} (b). We verified the effect of Learnable Weight Clipping on LLaMA-2-7B~\cite{touvron2023llama2}, and it can be observed that the weights $\mathbf{W}_{\mathbf{q}}$ of $q_{proj}$ in each layer have a smaller per-channel quantization MSE error compared to the original weights, as depicted in Fig.~\ref{fig:layerwise-mse} (c).

%%%%%%%%%%%%%%%%
% THE FOLLOWING TABLES MOVED QUITE AHEAD TO FIT PAGES
%%%%%%%%%%%%%%%%
\begin{table*}[ht]
\centering
\small
\setlength{\tabcolsep}{2pt} 
\begin{tabular}{ll*{7}{c}}
\toprule
Dataset & Method/LLaMA	&	Bits	&	LLaMA-7B	&	LLaMA-13B	&	LLaMA-65B	&	LLaMA-2-7B	&	LLaMA-2-13B	&	LLaMA-2-70B\\
\midrule
\multirow{5}{*}{LAMBADA}	&	FP16	&	W16A16	&	73.74\%	&	76.19\%	&	79.20\%	&	73.70\%	&	76.64\%	&	79.57\%\\
	&	AWQ-g128	&	W4A16	&	66.80\%	&	73.72\%	&	78.73\%	&	70.23\%	&	75.80\%	&	78.40\% \\
	&	GPTQ-g128	&	W4A16	&	70.21\%	&	75.68\%	&	78.77\%	&	72.31\%	&	75.99\%	&	79.86\% \\
	&	SmoothQuant$^*$	&	W8A8	&	73.49\%	&	76.15\%	&	78.07\%	&	73.36\%	&	76.05\%	&	78.71\%	\\
	&	OdysseyLLM	&	W4A8	&	73.49\%	&	76.23\%	&	78.56\%	&	70.81\%	&	76.07\%	&	79.43\% \\
\midrule
\multirow{5}{*}{C4}	&	FP16	&	W16A16	&	7.05	&	6.61	&	5.59	&	7.05	&	6.46	&	5.52 \\
	&	AWQ-g128	&	W4A16	&	7.34	&	6.81	&	5.71	&		7.42	&	6.68	&	5.63 \\
	&	GPTQ-g128	&	W4A16	&	7.71	&	6.73	&	5.70	&		8.74	&	6.6	&	5.60 \\
	&	SmoothQuant$^*$	&	W8A8	&	7.23	&	6.723	&	5.81	&	7.24	&	6.55	&	5.61 \\
	&	OdysseyLLM	&	W4A8	&	7.5	&	6.88	&	5.93	&	7.58	&	6.7	&	5.78 \\
\midrule
\multirow{5}{*}{WikiText}	&	FP16	&	W16A16	&	5.73	&	5.1	&	3.51	&	5.65	&	4.95	&	3.36\\
	&	AWQ-g128	&	W4A16	&	6.01	&	5.32	&	3.69	&	6.04	&	5.16	&	3.53\\
	&	GPTQ-g128	&	W4A16	&	6.31	&	5.24	&	3.67 &	6.36	&	5.1	&	3.50 \\
	&	SmoothQuant$^*$	&	W8A8	&	5.89	&	5.21	&	3.73	&	5.797	&	5.04	&	3.46\\
	&	OdesseyLLM	&	W4A8	&	6.17	&	5.37	&	3.92	&	6.11	&	5.19	&	3.7 \\
\bottomrule
 \end{tabular}
 \caption{Performance comparison for various quantized LLaMA models on LAMBADA, C4 and WikiText datasets. $^*$: per token for activations and per channel for weights}
 \label{tab:lambada-c5-wikitext}
\end{table*}

\begin{table*}[ht]
\centering
%\small
%\setlength{\tabcolsep}{2pt} 
\begin{tabular}{l*{7}{c}}
\toprule
Model & Method	&	BitWidth	&	WinoGrande	&	PIQA	&	HellaSwag	&	ARC\_e	&	Avg \\
\midrule
\multirow{5}{*}{LLaMA-1-7B} & FP16	&	W16A16	&	0.6985	&	0.7916	&	0.761	&	0.728	&	0.7448	\\
& AWQ-g128	&	W4A16	&	0.6938	&	0.7845	&	0.7465	&	0.7168	&	0.7354	\\
& GPTQ-g128	&	W4A16	&	0.6661	&	0.7786	&	0.7229	&	0.6557	&	0.7058	\\
& SmoothQuant$^*$	&	W8A8	&	0.7119	&	0.7894	&	0.7537	&	0.7386	&	0.7484	\\
& OdysseyLLM	&	W4A8	&	0.6977	&	0.7878	&	0.7407	&	0.7155	&	0.7354	\\
\midrule
\multirow{5}{*}{LLaMA-1-13B} & FP16	&	W16A16	&	0.7277	&	0.8009	&	0.7907	&	0.7471	&	0.7666	\\
& AWQ-g128	&	W4A16	&	0.7151	&	0.7971	&	0.7818	&	0.7256	&	0.7549	\\
& GPTQ-g128	&	W4A16	&	0.7238	&	0.8025	&	0.7828	&	0.7353	&	0.7611	\\
& SmoothQuant$^*$	&	W8A8	&	0.7238	&	0.802	&	0.7836	&	0.7466	&	0.764	\\
& OdysseyLLM	&	W4A8	&	0.7238	&	0.7998	&	0.7792	&	0.7441	&	0.7617	\\
\midrule
\multirow{5}{*}{LLaMA-1-65B} & FP16	&	W16A16	&	0.7735	&	0.8232	&	0.8415	&	0.7976	&	0.809	\\
& AWQ-g128	&	W4A16	&	0.7664	&	0.821	&	0.8427	&	0.7992	&	0.8073	\\
& GPTQ-g128	&	W4A16	&	0.7632	&	0.8221	&	0.8382	&	0.79	&	0.8034	\\
& SmoothQuant$^*$	&	W8A8	&	0.7593	&	0.7976	&	0.8097	&	0.7471	&	0.7784	\\
& OdysseyLLM	&	W4A8	&	0.753	&	0.7933	&	0.8019	&	0.734	&	0.7706	\\
\midrule
\multirow{5}{*}{LLaMA-2-7B} & FP16	&	W16A16	&	0.6906	&	0.7911	&	0.7598	&	0.7458	&	0.7468	\\
& AWQ-g128	&	W4A16	&	0.6819	&	0.7786	&	0.7473	&	0.6675	&	0.7188	\\
& GPTQ-g128	&	W4A16	&	0.6772	&	0.7845	&	0.748	&	0.6742	&	0.721	\\
& SmoothQuant$^*$	&	W8A8	&	0.6875	&	0.7873	&	0.7598	&	0.7104	&	0.7363	\\
& OdysseyLLM	&	W4A8	&	0.6811	&	0.7742	&	0.7398	&	0.6953	&	0.7226	\\
\midrule
\multirow{5}{*}{LLaMA-2-13B} & FP16	&	W16A16	&	0.7222	&	0.8052	&	0.7938	&	0.7744	&	0.7739	\\
& AWQ-g128	&	W4A16	&	0.7253	&	0.7987	&	0.7838	&	0.766	&	0.7685	\\
& GPTQ-g128	&	W4A16	&	0.7245	&	0.7992	&	0.7899	&	0.7736	&	0.7718	\\
& SmoothQuant$^*$	&	W8A8	&	0.723	&	0.8052	&	0.7977	&	0.7681	&	0.7735	\\
& OdysseyLLM	&	W4A8	&	0.7111	&	0.7976	&	0.7782	&	0.763	&	0.7625	\\
\midrule
\multirow{5}{*}{LLaMA-2-70B} & FP16	&	W16A16	&	0.7798	&	0.8275	&	0.8381	&	0.8098	&	0.8138	\\
& AWQ-g128	&	W4A16	&	0.7727	&	0.8281	&	0.8341	&	0.803	&	0.8095	\\
& GPTQ-g128	&	W4A16	&	0.779	&	0.833	&	0.8343	&	0.8035	&	0.8125	\\
& SmoothQuant$^*$	&	W8A8	&	0.7766	&	0.8303	&	0.8345	&	0.8127	&	0.8135	\\
& OdysseyLLM	&	W4A8	&	0.7751	&	0.8313	&	0.8272	&	0.806	&	0.8099 \\
\bottomrule
 \end{tabular}
 \caption{Comparison on Common Sense QA. $^*$: per token for activations and per channel for weights}
 \label{tab:common-sense-qa}
\end{table*}

%%%%%%%%%%%%%%%%%%%%%%%%%%%%%%

\subsection{Hessian-based Training-free Compensation}\label{subsec:hessian}
Although \emph{learnable weight clipping} can effectively alleviate the loss brought by 4-bit integer per-channel quantization, further compensation is still required to enhance the performance. We generally follow a layerwise quantization structure~\cite{nagel2020up,li2021brecq} that reduces the mean square error before and after quantization. In particular, we choose GPTQ~\cite{frantar2022gptq} that speeds up the Hessian-based quantization compensation algorithm OBQ~\cite{frantar2022obc} by parallel execution and the removal of greedy strategy. Essentially, this genre of algorithms iteratively updates the remaining set $F$ of full-precision weight with $ \boldsymbol{\delta}_F$ to offset the error brought by the quantized weight $\mathbf{Q} (\mathbf{W}_i)$, as formulated by Eq.~\ref{eq:hessian}, where $\mathbf{W}_i$ denotes the weights in the $i$-th row and $\mathbf{H}_F = 2 \mathbf{X}_F \mathbf{X}_F^{\top}$.

\begin{align}\label{eq:hessian}
\mathbf{W}_i&=\operatorname{argmin}_{\mathbf{W}_i} \frac{\left( \mathbf{Q} \left(\mathbf{W}_i\right)-\mathbf{W}_i\right)^2}{\left[\mathbf{H}_F^{-1}\right]_{ii}} \\
 \boldsymbol{\delta}_F&=-\frac{\mathbf{W}_i- \mathbf{Q} \left(\mathbf{W}_i\right)}{\left[\mathbf{H}_F^{-1}\right]_{ii}} \cdot\left(\mathbf{H}_F^{-1}\right)_{:, i}
\end{align}

\subsection{Fast Mixed-Precision GEMM}\label{subsec:gemm}

The W4A8 calls for a renovation for the kernel implementation to maximize the benefit of this bit width setting on mainstream hardware. We take the following three steps to obtain FastGEMM.

\begin{figure}[ht]
\centering
\includegraphics[width=0.5\textwidth]{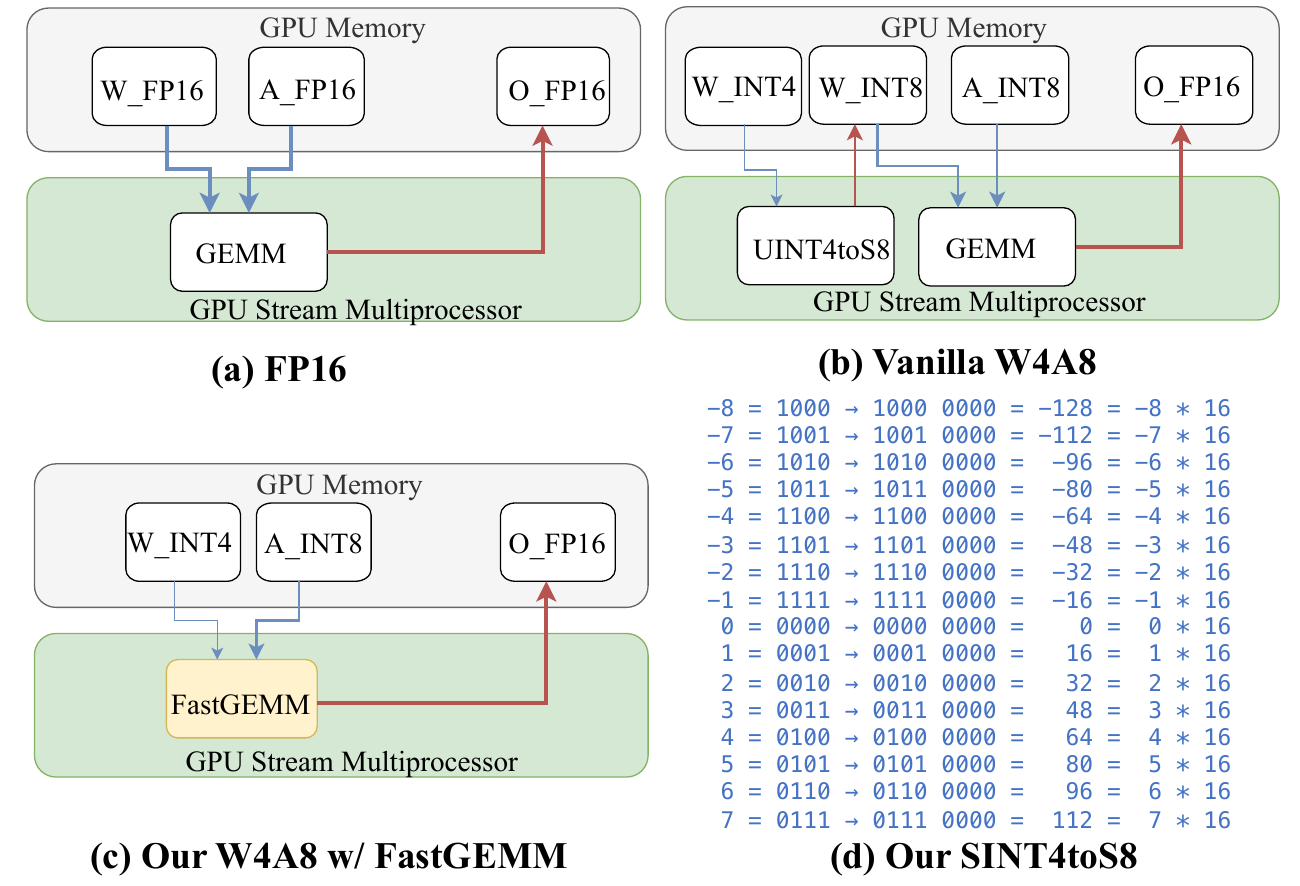}
\caption{Design of W4A8 FastGEMM (c) compared to FP16 (a) and vanilla version (b). (d) Our \texttt{SInt4to8} conversion (represented in two's complement) where 4-bit integers are multiplied by 16 to have a convenient 8-bit GEMM.}
\label{fig:fastgemm}
\end{figure}

\textbf{Kernel fusion.} As shown in Fig.~\ref{fig:fastgemm} (a), modern GPUs only support GEMM calculations of the same type, while mixed precision requires conversion to the same type first. Depicted in Fig.~\ref{fig:fastgemm} (b), a naive approach is to implement a separate GPU Kernel to perform type conversion, but this would increase additional memory access and substantially slow down the model inference speed. Instead, we propose to fuse \texttt{SINT4toS8} and GEMM into one GPU kernel, abbreviated as FastGEMM and shown in Fig.~\ref{fig:fastgemm} (c). However, it is non-trivial to achieve.

\noindent \textbf{Removal of INT8 subtraction.} As mentioned in Sec.~\ref{subsec:adapt}, we adopt symmetric quantization. This is beneficial for two reasons. First, according to LLM-QAT~\cite{liu2023llm}, the LLaMA series performs better in terms of accuracy with symmetric quantization compared to asymmetric quantization. Second, modern GPUs like NVIDIA, in order to reduce the size of the chip, do not provide subtraction instructions for the signed 8-bit integers\footnote{\href{https://goo.by/RjASnb}{https://docs.nvidia.com/cuda/parallel-thread-execution/index.html\#integer-arithmetic-instructions-sub}}. This leads to additional type conversion instructions when asymmetric quantization is processing the zero point, thereby impeding the inference speed. Choosing symmetric quantization directly removes the zero-point subtraction.

\begin{figure}[ht]
\centering
\includegraphics[width=0.45\textwidth]{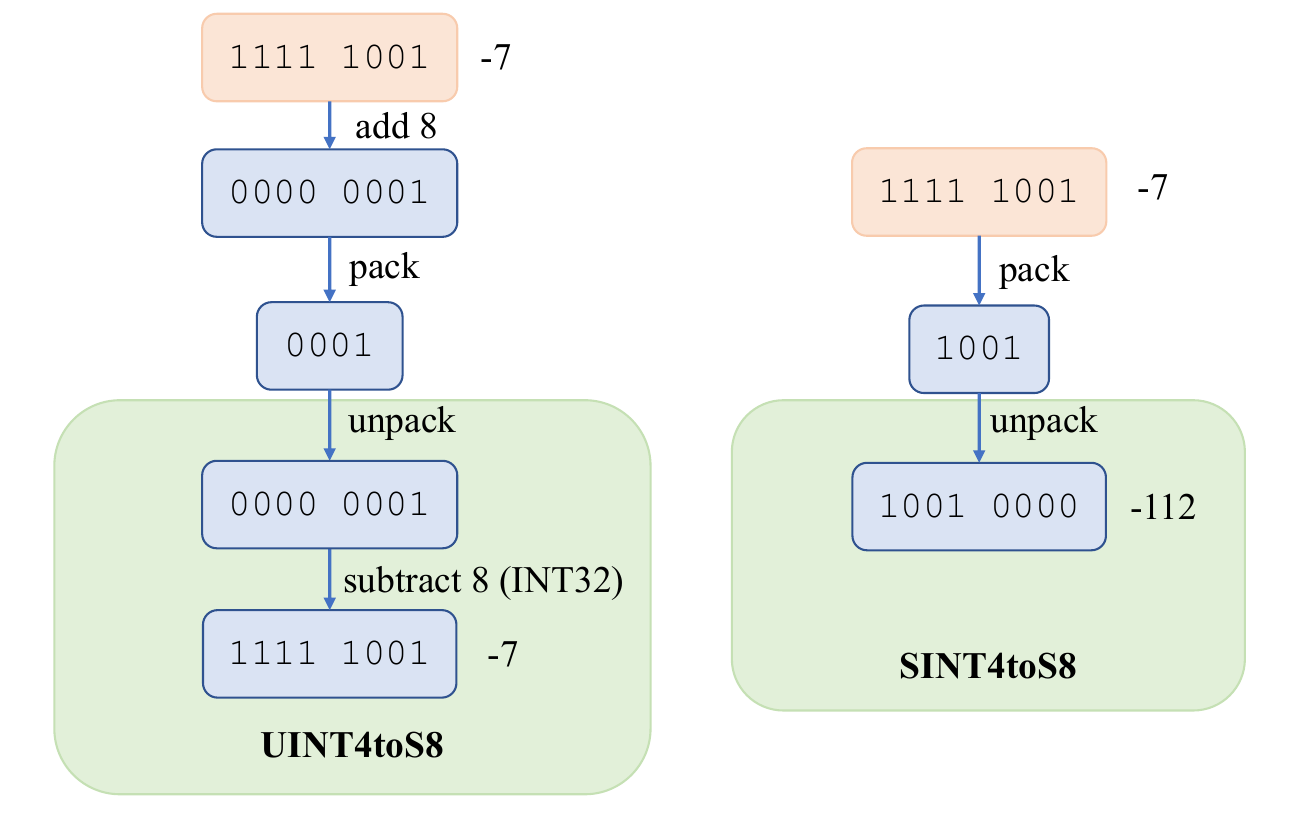}
\caption{Vanilla W4A8's \texttt{UINT4toS8} vs. our proposed \texttt{SINT4toS8}. The green area is on the device, the rest is offline.}
\label{fig:conversion}
\end{figure}

\noindent \textbf{Reusing the sign bit.} In Fig.~\ref{fig:conversion}, a vanilla W4A8's \texttt{UINT4toS8} operation can be very costly. For instance, to load the two's complement of -7 on GPU, it requires sophisticated calculation (detailed in Sec.~\ref{app:sub:conversion} (supp.)), especially there is an additional on-device subtraction which is not directly supported by hardware. By default, it has to be converted to higher precision (typically INT32) for such subtraction which incurs substantial cost. We implement it as Asym GEMM in Fig.~\ref{fig:fastgemm-comp} to see how costly it is. As we are motivated to explore the authentic benefit of W4A8, we have to fabricate a faster W4A8 scheme to load signed INT4 to signed INT8. Specifically, we store the weights of signed INT4 into two's complement, shown in Fig.~\ref{fig:fastgemm} (d). During computation, we place the signed INT4 weights in the higher 4 bits of signed INT8, which is equivalent to multiplying each value by 16. After the completion of the GEMM calculation, we divide the output value by 16 to restore the correct results. Such multiplication and division can be easily and efficiently implemented. Since the internal accumulator is of the INT32 type, there will be no overflow. With this novel conversion scheme, we can retain accuracy while significantly improving inference speed and can be used out-of-box on modern GPUs.

%%%%%%%%%%%% MOVED FOR LAYOUT %%%%%%%%%%%%%%
\begin{figure*}[ht]
\centering
\includegraphics[width=0.9\textwidth]{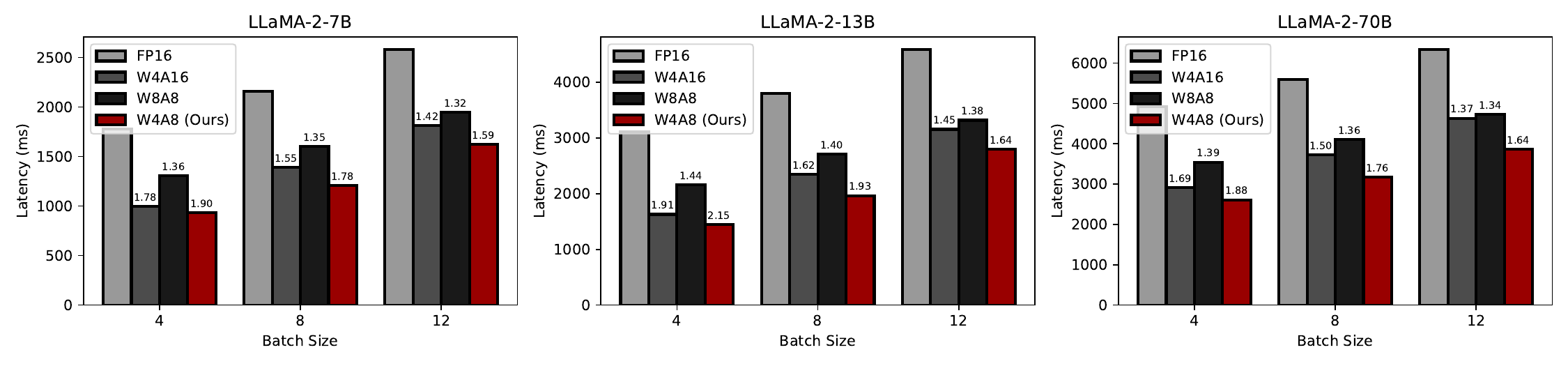}
\caption{Latency comparison on LLaMA-2 models in various bitwidth settings.}
\label{fig:latency-comp}
%\vskip -0.1in
\end{figure*}
%%%%%%%%%%%%%%%%%%%%%%%%%%%%%%%%%%%%%%

\section{Experiments}\label{sec:exp}

\subsection{Settings}\label{subsec:setup}
\paragraph{Inference Implementation}
The latency comparison in Fig.~\ref{fig:perf} and Fig.~\ref{fig:latency-comp} is made fair by utilizing the same set of configurations except for bit widths. Specifically, we implemented the whole end-to-end inference pipeline with CUTLASS~\cite{cutlass} to have a delicate combination of GPU Tensor Core execution, kernel fusion policy, and graph optimization. We also evaluate the performance with the latest build of TensorRT-LLM~\cite{tensorrt-llm}. All latency measurements in the paper are tested on NVIDIA A100 80G GPUs.

\paragraph{Models and Datasets} 
Following recent quantization methods ~\cite{liu2023llmqat,frantar2022gptq}, we evaluate our approach by conducting experiments on LLaMA series~\cite{touvron2023llama,touvron2023llama2}
models and presenting results on various tasks. We randomly pick 128 sequences in C4 ~\cite{C4} datasets for calibration. We report the zero-shot performance on Common Sense Reasoning tasks such as PIQA \cite{bisk2020piqa}, HellaSwag ~\cite{zellers2019hellaswag}, WinoGrande~\cite{sakaguchi2021winogrande}, ARC ~\cite{clark2018think}. We also assess the few-shot performance on Common Sense~\cite{talmor2019commonsenseqa} and MMLU ~\cite{hendrycks2020measuring} datasets, along with perplexity
scores on WikiText2 ~\cite{wikitext103} and C4 ~\cite{C4} datasets.

\subsection{Performance Comparison}

Table~\ref{tab:lambada-c5-wikitext} presents the performance comparison on various common datasets LAMBADA, C4, and WikiText, while Table~\ref{tab:common-sense-qa} is on Common Sense QA and Table~\ref{tab:mmlu-qa} (supp.) on MMLU. We compare our OdysseyLLM with the most recent state-of-art quantization methods AWQ~\cite{lin2023awq}, SmoothQuant~\cite{xiao2023smoothquant}, and GPTQ~\cite{frantar2022gptq}. It turns out that our W4A8 OdysseyLLM mostly achieves on-par performance with the state-of-the-art W8A8 approach SmoothQuant on a large range of tasks, paving the way to its ready application in the real world. Being a post-training quantization method, we also enjoy the low-cost benefit during the quantization process. 

\subsection{Latency Comparison on LLaMA models}

We exhibit the overall latency comparison of LLaMA-2 models under the same implementations (Sec.~\ref{subsec:setup}) except the bandwidths in Fig.~\ref{fig:latency-comp} where our W4A8 version prevails on all model scales. Notably, we achieve at most 1.9$\times$, 2.15$\times$, and 1.76$\times$ boost compared with FP16 for LLaMA-2 7B, 13B, and 70B respectively. We use 1 GPU for 7B, and 13B, 4 GPUs for 70B. All inputs have an input sequence length of 1024. Output tokens are set to 128.

\subsection{Comparison with TensorRT-LLM}
TensorRT-LLM~\cite{tensorrt-llm} is so far the most advanced industry-level deployment engine for LLMs, shipped with both FP16 and INT8 implementation. In Table~\ref{tab:tensorrt-llm}, we compare with TensorRT-LLM to show the benefits of our inference engine and newly fabricated kernel. The settings are kept the same as in Fig.~\ref{fig:latency-comp} except that here we use a batch size of 1. Notice that our engine is mostly comparable to TensorRT-LLM in both FP16 and W8A8 settings. Our engine with the new W4A8 kernel obtains 1.37$\times$, 1.45$\times$, and 1.36$\times$ boosts for LLaMA-2 models against the W8A8 setting in TensorRT-LLM (no available W4A8 yet), also 1.87$\times$, 2.23$\times$, and 1.83$\times$ against its FP16 setting.

\begin{table}[ht!]
\centering
\small
\setlength{\tabcolsep}{2pt} 
\begin{tabular}{l|*{2}{c}|*{3}{c}}
\toprule
Model & \multicolumn{2}{c|}{TensorRT-LLM}  &    \multicolumn{3}{c}{Ours}   \\
      & FP16 & W8A8 & FP16 & W8A8 & W4A8 \\
\midrule
LLaMA-2-7B  & 1411 & 1030 & 1513 & 1103 & 751 (1.37$\times$)\\	
LLaMA-2-13B & 2547 & 1657 & 2671 & 1824 & 1139 (1.45$\times$) \\	
LLaMA-2-70B & 4177 & 3087 & 4271 & 3135 & 2263 (1.36$\times$)  \\	
\bottomrule
 \end{tabular}
 \caption{Latency comparison (in $ms$) with TensorRT-LLM.} %Our W4A8 boosts is against the W8A8 of TensorRT-LLM.
 \label{tab:tensorrt-llm}
\end{table}

\subsection{Comparison with QUIK}

QUIK~\cite{ashkboos2023towards} comes with a W4A4 implementation while outliers fall back to higher precision. We show that how such an approach renders an overall inferior speed in practice. Per-kernel measurements are shown in Table~\ref{tab:quik} where our speed can be 4.33$\times$ faster in the self-decoding stage. QUIK is only on par with our speed at the context decoding stage since it is more computation-intensive. This benefit is quickly amortized throughout an end-to-end setting. See Sec.~\ref{app:sec:quik-analysis} for a detailed analysis.

\begin{table}[ht!]
\centering
\small
\setlength{\tabcolsep}{2pt} 
\begin{tabular}{l*{6}{c}}
\toprule
Stage & M & N & K & QUIK & Odyssey & Boost \\
\midrule
\multirow{4}{*}{Context decode} & \multirow{4}{*}{1024}	&	4096 	&	4096	 &	0.139	&	0.121	&	1.14$\times$ \\
	 					&	 		 		 	& 1024	&	8192	&	0.095	&	0.073	&	\textbf{1.30$\times$} \\
						&	 		 		 	& 11088	&	4096	&	0.290	&	0.279	&	1.03$\times$ \\
						&	 		 		 	& 5120	&	5120	&	0.163	&	0.158	&	1.03$\times$ \\
\midrule
\multirow{4}{*}{Self-decode} & \multirow{4}{*}{1}	 	& 4096 	&	4096	&	0.052	&	0.012	&	\textbf{4.33$\times$} \\
					&	 		 		 	& 1024	&	8192	&	0.080	&	0.019	&	4.21$\times$ \\
					&	 		 		 	& 11088	&	4096	&	0.054	&	0.016	&	3.37$\times$ \\
					& 	 		 		 	& 5120	&	5120	&	0.060	&	0.014 	&	4.28$\times$ \\
\bottomrule
 \end{tabular}
 \caption{GEMM latency comparison with QUIK. N stands for the output dimension of weight, M$\times$K for activation shape}
 \label{tab:quik}
\end{table}

\section{Ablation Study}\label{sec:ablation}

\subsection{Quantization Strategy}

Table~\ref{tab:abl-quant-strategy} justifies our choices of symmetric LWC and GPTQ. Vanilla W4A8 which doesn't involve compensation techniques falls short in the performance (PPL) on WikiText2 and C4. The recipe of LWC and GPTQ combined generally produces the best result.
%Norm Tweaking~\cite{li2023norm} could be an orthogonal compensation strategy. 

\begin{table}[ht!]
\centering
\small
\setlength{\tabcolsep}{1pt} 
\begin{tabular}{l*{4}{|c}}
\toprule
Dataset & Model & Baseline & B+LWC & B+LWC+GPTQ \\
\midrule
\multirow{6}{*}{WikiText2} & LLaMA-1-7B & 6.73 & 6.25 & \textbf{6.17}  \\
 & LLaMA-1-13B & 5.7 & \textbf{5.37} & \textbf{5.37} \\
 & LLaMA-1-65B & 4.41 & \textbf{3.89} & 3.92 \\
 & LLaMA-2-7B & 7.13 & 6.73 & \textbf{6.11} \\
 & LLaMA-2-13B & 5.47 & 5.30 & \textbf{5.19} \\
 & LLaMA-2-70B & 3.93 & 3.74 & \textbf{3.70}  \\
\midrule
\multirow{6}{*}{C4} & LLaMA-1-7B & 8.16 & 7.64 & \textbf{7.50} \\
 & LLaMA-1-13B & 7.24 & 6.92 & \textbf{6.88} \\
 & LLaMA-1-65B & 6.35 & 5.97 & \textbf{5.93} \\
 & LLaMA-2-7B & 8.88 & 8.54 & \textbf{7.58} \\
 & LLaMA-2-13B & 7.0 & 6.84 & \textbf{6.70} \\
 & LLaMA-2-70B & 6.01 & 5.83 & \textbf{5.78} \\
\bottomrule
 \end{tabular}
 \caption{PPL on WikiText2 and C4. Baseline (B): Vanilla W4A8, LWC: symmetric learnable weight clipping}
 \label{tab:abl-quant-strategy}
\end{table}

\subsection{FastGEMM vs. Fine-grained vs. Asymmetric}

\begin{figure}[ht]
\centering
\includegraphics[width=0.45\textwidth]{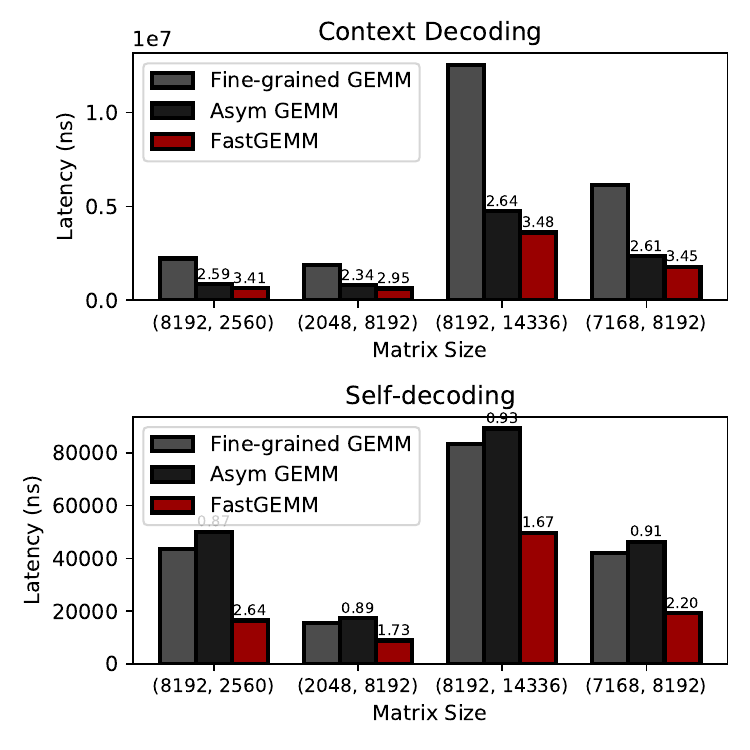}
\caption{Latency comparison (measured in nanoseconds) of all GEMMs in LLaMA-2-70B with tensor parallelism of 4 at two decoding stages. The matrix size is the input and output dimensions of GEMM. We use a batch size of 8 and an input length of 1024. The number atop the bar is the boost w.r.t. fine-grained GEMM.}
%\vskip -0.1in
\label{fig:fastgemm-comp}
\end{figure}

We conducted an in-depth study on different matrix multiplication (GEMM) implementation strategies on the LLaMA-2-70B model to understand their impact on performance. As shown in Fig.~\ref{fig:fastgemm-comp}, the horizontal axis represents the GEMM Size $(dim_{i}, dim_{o})$ of the model under a partitioning on 4 GPUs, with an input length of 1024 and a batch size of 8. Noticeably, fine-grained GEMM requires frequent dequantization operations per group, introducing a large amount of \texttt{Integer2Float} and Fused Multiply-Add (FMA) overhead; the signed 8-bit subtraction operation in Asymmetric GEMM needs to fallback to signed 32-bit, introducing additional conversion cost. In contrast, our FastGEMM well solves the drawbacks of these two GEMMs, achieving the best performance.

\section{Conclusion}\label{sec:concl}

Under the hardware-centric guidance, our paper introduces the first deployable W4A8 solution for LLMs. We give a composition of recipes named OdysseyLLM which comprises symmetric learnable weight clipping, iterative Hessian-based compensation, and a novel FastGEMM for an accelerated W4A8 calculation. To our knowledge, we have achieved the fastest W4A8 LLMs available so far, with acceptable quantization loss on common language benchmarks. We believe this exploration can serve as a solid ground for further readily applicable compression algorithms to emerge, reducing LLM inference cost and facilitating potential applications in constrained scenarios.

\clearpage
\clearpage
{
  \small
  \bibliographystyle{ieeenat_fullname}
  \bibliography{main}
}


\appendix
\newpage
\newpage
\section{Kernel Implementation}

\subsection{UINT4toS8 vs. SINT4toS8}\label{app:sub:conversion}

In modern computer systems, integer numbers are commonly stored as two's complement for multiple known benefits. To move these numbers from host to device requires specific offline preprocessing. As illustrated in Fig.~\ref{fig:conversion} (main text), say we have a signed integer -7 represented in its two's complement as \texttt{1111 1001}. We first have to rearrange within the range (0,15) by adding 8 to have an unsigned integer \texttt{0000 0001}. The lower 4-bit \texttt{0001} can be utilized for weight packing (\eg. four UINT4 integers packed in 32 bits). Once the GPU receives such 32 bits, it is unpacked first to obtain each of these 4 numbers. Here comes the problem, to revert such a UINT4 number to an SINT8 for later GEMM computation, we have to subtract 8 again, which is not directly available for GPUs. One has to convert it to INT32 to enable subtraction, which is very costly in practice. To avoid the unexpected cost, we simplify the pipeline by directly packing the lower 4 bits \texttt{1001}. During unpacking, we place them in the higher 4 bits on a piece of INT8 GPU memory, which in effect renders a signed integer number in 8 bits, but 16 times larger. We later divide the GEMM result by 16 to restore the correct value. This new implementation substantially eases the pain of type conversion and speeds up the overall performance of the FastGEMM kernel.

\subsection{Analysis on QUIK's Latency}\label{app:sec:quik-analysis}

In Table~\ref{tab:quik} in the main text, it has been observed that QUIK's performance is substantially poor during the self-decoding phase. We discover the reason lies in their various separated CUTLUSS kernels to adapt the mixed precision recipe. Ideally, pure W4A4 computation would be 2$\times$ faster than W4A8. However, during the memory-bound self-decoding phase, theoretical I/O overhead between W4A4 and W4A8 is nearly the same since the sequence length of activation becomes very small (typically one token at a time). Thus the benefit of W4A4 vanishes. In QUIK, their aggregated I/O overhead on various kernels becomes quite significant, which deteriorates its performance. While our FastGEMM enjoys kernel fusion, single kernel design leads to at most \textbf{4.33$\times$} speed boosting.

\subsection{Latency Comparison with Hugging Face}

Hugging Face\footnote{\url{https://huggingface.co/blog/4bit-transformers-bitsandbytes}} provides a 4-bit implementation with the \texttt{bitsandbytes} library \footnote{\url{https://github.com/TimDettmers/bitsandbytes}}. We compare their latencies in Table~\ref{tab:comp-hf-4bit}. Note that this 4-bit implementation is even slower than Hugging Face's FP16 implementation, which prevents it from being a real application. It adopts a particular normal format 4-bit (NF4)~\cite{dettmers20218} to pursue a higher precision and reduced memory, however, it comes at the cost of an extremely complex computation strategy which ultimately leads to even worse speed compared with FP16.

\begin{table}[ht!]
\centering
\small
\setlength{\tabcolsep}{1pt} 
\begin{tabular}{l|c|*{2}{c}|*{3}{c}}
\toprule
Model & BS & \multicolumn{2}{c|}{Hugging Face}  &    \multicolumn{3}{c}{Ours}   \\
      & & FP16 & 4-bit & W4A8 & vs. HF F16 & vs. HF 4-bit \\
\midrule
LLaMA-2-7B   & 1 & 3439 & 6602 & 751 & 4.57$\times$ & 8.78$\times$ \\		
LLaMA-2-7B   & 4 & 3769 & 10790 & 935 & 4.03$\times$ & 11.53$\times$ \\	
LLaMA-2-13B  & 1 & 4578 & 8596 & 1139 & 4.01$\times$ & 7.54$\times$\\	
LLaMA-2-13B  & 4 & 5610 & 19435 & 1447 & 3.87$\times$ & 13.42$\times$\\	
\bottomrule
 \end{tabular}
 \caption{Latency comparison (in $ms$) with Hugging Face.} %Our W4A8 boosts is against the W8A8 of TensorRT-LLM.
 \label{tab:comp-hf-4bit}
\end{table}

\section{Additional Experiments}
\subsection{Comparison on MMLU}

Table.~\ref{tab:mmlu-qa} gives OdysseyLLM compared with the state-of-the-art methods on MMLU, which are mostly comparable with the W4A8 solution.

\begin{table*}[ht]
\centering
\small
\begin{tabular}{l*{7}{c}}
\toprule
Model	&	Method	&	BitWidth	&	Hums.	&	STEM	&	Social	&	Other	&	Avg\\
\midrule
\multirow{5}{*}{LLaMa-1-7B}	&	FP16	&	W16A16	&	33.65\%	&	31.05\%	&	38.22\%	&	38.43\%	&	35.19\%\\
	&	AWQ-g128	&	W4A16	&	31.86\%	&	30.62\%	&	35.72\%	&	38.62\%	&	34.00\%\\
	&	GPTQ-g128	&	W4A16	&	31.56\%	&	29.82\%	&	36.69\%	&	36.34\%	&	33.41\%\\
	&	SmoothQuant$^*$	&	W8A8	&	33.90\%	&	30.75\%	&	37.80\%	&	40.19\%	&	35.53\%\\
	&	OdysseyLLM	&	W4A8	&	32.56\%	&	30.38\%	&	35.13\%	&	38.49\%	&	34.03\%\\
\midrule
\multirow{5}{*}{LLaMa-1-13B}	&	FP16	&	W16A16	&	44.61\%	&	37.08\%	&	54.05\%	&	53.52\%	&	47.12\%\\
	&	AWQ-g128	&	W4A16	&	43.23\%	&	34.86\%	&	51.41\%	&	51.20\%	&	45.06\%\\
	&	GPTQ-g128	&	W4A16	&	42.98\%	&	36.28\%	&	52.42\%	&	51.76\%	&	45.63\%\\
	&	SmoothQuant$^*$	&	W8A8	&	44.25\%	&	35.98\%	&	52.97\%	&	52.38\%	&	46.26\%\\
	&	OdysseyLLM	&	W4A8	&	42.15\%	&	35.69\%	&	51.48\%	&	50.74\%	&	44.79\%\\
\midrule
\multirow{5}{*}{LLaMa-1-65B}	&	FP16	&	W16A16	&	61.76\%	&	51.99\%	&	73.29\%	&	67.58\%	&	63.53\%\\
	&	AWQ-g128	&	W4A16	&	60.66\%	&	50.93\%	&	71.53\%	&	66.47\%	&	62.29\%\\
	&	GPTQ-g128	&	W4A16	&	60.40\%	&	51.16\%	&	71.66\%	&	66.72\%	&	62.34\%\\
	&	SmoothQuant$^*$	&	W8A8	&	61.23\%	&	51.06\%	&	71.73\%	&	67.27\%	&	62.74\%\\
	&	OdysseyLLM	&	W4A8	&	59.72\%	&	48.64\%	&	71.56\%	&	65.76\%	&	61.33\%\\
\midrule
\multirow{5}{*}{LLaMa-2-7B}	&	FP16	&	W16A16	&	36.92\%	&	30.75\%	&	40.92\%	&	45.68\%	&	38.49\%\\
	&	AWQ-g128	&	W4A16	&	32.62\%	&	31.64\%	&	39.71\%	&	42.66\%	&	36.28\%\\
	&	GPTQ-g128	&	W4A16	&	36.20\%	&	31.88\%	&	40.23\%	&	44.51\%	&	38.07\%\\
	&	SmoothQuant$^*$	&	W8A8	&	34.77\%	&	29.62\%	&	38.58\%	&	43.46\%	&	36.50\%\\
	&	OdysseyLLM	&	W4A8	&	34.41\%	&	28.83\%	&	40.66\%	&	41.39\%	&	36.19\%\\
\midrule
\multirow{5}{*}{LLaMa-2-13B}	&	FP16	&	W16A16	&	54.43\%	&	44.27\%	&	63.41\%	&	60.76\%	&	55.68\%\\
	&	AWQ-g128	&	W4A16	&	50.63\%	&	42.21\%	&	62.01\%	&	59.22\%	&	53.30\%\\
	&	GPTQ-g128	&	W4A16	&	50.44\%	&	43.34\%	&	62.14\%	&	60.27\%	&	53.75\%\\
	&	SmoothQuant$^*$	&	W8A8	&	53.28\%	&	44.14\%	&	63.54\%	&	60.86\%	&	55.31\%\\
	&	OdysseyLLM	&	W4A8	&	50.78\%	&	42.41\%	&	61.13\%	&	59.04\%	&	53.15\%\\
\midrule
\multirow{5}{*}{LLaMa-2-70B}	&	FP16	&	W16A16	&	65.16\%	&	57.79\%	&	80.44\%	&	74.61\%	&	69.11\%\\
	&	AWQ-g128	&	W4A16	&	64.44\%	&	57.89\%	&	79.62\%	&	73.60\%	&	68.47\%\\
	&	GPTQ-g128	&	W4A16	&	64.02\%	&	56.66\%	&	80.11\%	&	74.06\%	&	68.28\%\\
	&	SmoothQuant$^*$	&	W8A8	&	63.53\%	&	56.00\%	&	79.23\%	&	73.81\%	&	67.73\%\\
	&	OdysseyLLM	&	W4A8	&	63.12\%	&	55.40\%	&	78.29\%	&	72.49\%	&	66.95\% \\
\bottomrule
 \end{tabular}
 \caption{Comparison on MMLU. $^*$: per token for activations and per channel for weights}
 \label{tab:mmlu-qa}
\end{table*}

\end{document}